\documentclass[letterpaper, 10 pt, conference]{ieeeconf}  

\usepackage{times}
\usepackage{epsfig}
\usepackage{graphicx}
\usepackage{amsmath}
\usepackage{amssymb}
\usepackage{listliketab}
\usepackage{tabto}
\usepackage[dvipsnames]{xcolor}
\usepackage{float}

\usepackage{paralist}

\usepackage{amsmath}
\usepackage{amssymb}
\usepackage{mathtools,xparse}
\usepackage{algorithm}
\usepackage[noend]{algpseudocode}
\usepackage{dsfont}
\usepackage{hyperref}

\makeatletter
\def\BState{\State\hskip-\ALG@thistlm}
\makeatother

\IEEEoverridecommandlockouts                              

\title{\LARGE \bf
Dexterous Humanoid Hand Control for Human-Robot Interactions Using Deep Reinforcement Learning and Imitation }

\title{\LARGE \bf
Demonstration-Guided Deep Reinforcement Learning of Control Policies for Dexterous Human-Robot Interaction}

\author{Sammy Christen$^{1}$, Stefan Stev{\v{s}}i{\'{c}}$^{1}$,  Otmar Hilliges$^{1}$
\thanks{$^1$AIT Lab, Department of Computer Science, ETH Zurich,   8092 Zurich, Switzerland        {\tt\small sammy.christen | stefan.stevsic | otmar.hilliges @inf.ethz.ch}}%
\thanks{ This work was supported in parts by the Swiss National Science Foundation (UFO 200021L\_153644).  We thank the NVIDIA Corporation for the donation of GPU servers used in this work.}
}

\begin{document}

\maketitle
\thispagestyle{empty}
\pagestyle{empty}


\begin{abstract}
In this paper, we propose a method for training control policies for human-robot interactions such as handshakes or hand claps via Deep Reinforcement Learning.
The policy controls a humanoid Shadow Dexterous Hand, attached to a robot arm. 
We propose a parameterizable multi-objective reward function that allows learning of a variety of interactions without changing the reward structure. 
The parameters of the reward function are estimated directly from motion capture data of human-human interactions in order to produce policies that are perceived as being natural and human-like by observers. 
We evaluate our method on three significantly different hand interactions: handshake, hand clap and finger touch. 
We provide detailed analysis of the proposed reward function and the resulting policies and conduct a large-scale user study, indicating that our policy produces natural looking motions.
\end{abstract}


\section{Introduction}

Dexterous humanoid hands, such as the Shadow Dexterous Hand \cite{shadowHand}, are becoming very sophisticated. Improvements in mechatronics have enabled very compact systems that have more than twenty degrees of freedom (DoF). However, controller design remains very challenging and has been shown to be a very complex problem \cite{rajeswaran2017learning, mordatch2012contact}. Recently, model-free deep reinforcement learning (DRL) algorithms have been applied to the control of humanoid hands, albeit on relatively simple tasks such as grasping or door opening \cite{rajeswaran2017learning} and in simulation only. 
In \cite{openAI2018}, a controller trained in simulation has been transferred to a real humanoid hand. This opens up the door to learn policies for natural physical human-robot interactions. In particular, we are interested in learning a control policy for diverse hand interactions, such as handshakes or hand claps. The handshake is the most common greeting gesture throughout the world, therefore it has received a lot of attention in the robotics community \cite{knoop2017handshakiness, orefice2016let, arns2017design, shu2017learning, jindai2015handshake}. In this paper, we present a method for training a control policy for human-robot hand interactions, using data from human demonstrations in combination with deep reinforcement learning. We test our method on the simulated model of the Shadow Dexterous Hand.

To train a control policy using a DRL algorithm, one of the main issues is the definition of a reward function. For simple tasks, like grasping or pick and place tasks, the goal is obvious and the reward can be easily shaped. For our task, however, it is not obvious how to shape a reward function. The reward needs to result in motions that are perceived as natural, the hand needs to reach a desired contact profile and precise position, while dealing with complex contact dynamics.
To produce natural looking motions of animated characters in \cite{peng2018deepmimic}, the reward  is based on tracking position and angle references from motion capture data. However, the authors only consider motions in open space and hence their reward function cannot be transfered to our task. 
Thus, we investigate important terms to construct a reward function and compare the influence of different reward terms in an ablation study.  
To enable generalization to different hand interactions, we define a parametrized reward function. 
We extract most of the reward function parameters from motion capture data, leading to only six parameters that are relatively easy to adjust. 
One could argue that the policy could be learned directly from data via Inverse Reinforcement Learning (IRL). However, state-of-the-art IRL methods \cite{ho2016generative, merel2017learning} have not been applied to tasks that require precise positioning or challenging contact dynamics. Furthermore, these methods can be unstable when applied to motion capture data \cite{merel2017learning}.  
To ensure training convergence, we propose a specialized training method. Standard DRL algorithms work out of the box on benchmark problems \cite{openAiGym}, but for more complex problems additional training details, such as randomization or early stopping are important \cite{peng2018deepmimic, vecerik2017leveraging, rajeswaran2017learning}.
We propose a training method which works in combination with DDPG, resulting in stable convergence properties.

\begin{figure}
\centering
	\includegraphics[width= 0.85 \linewidth]{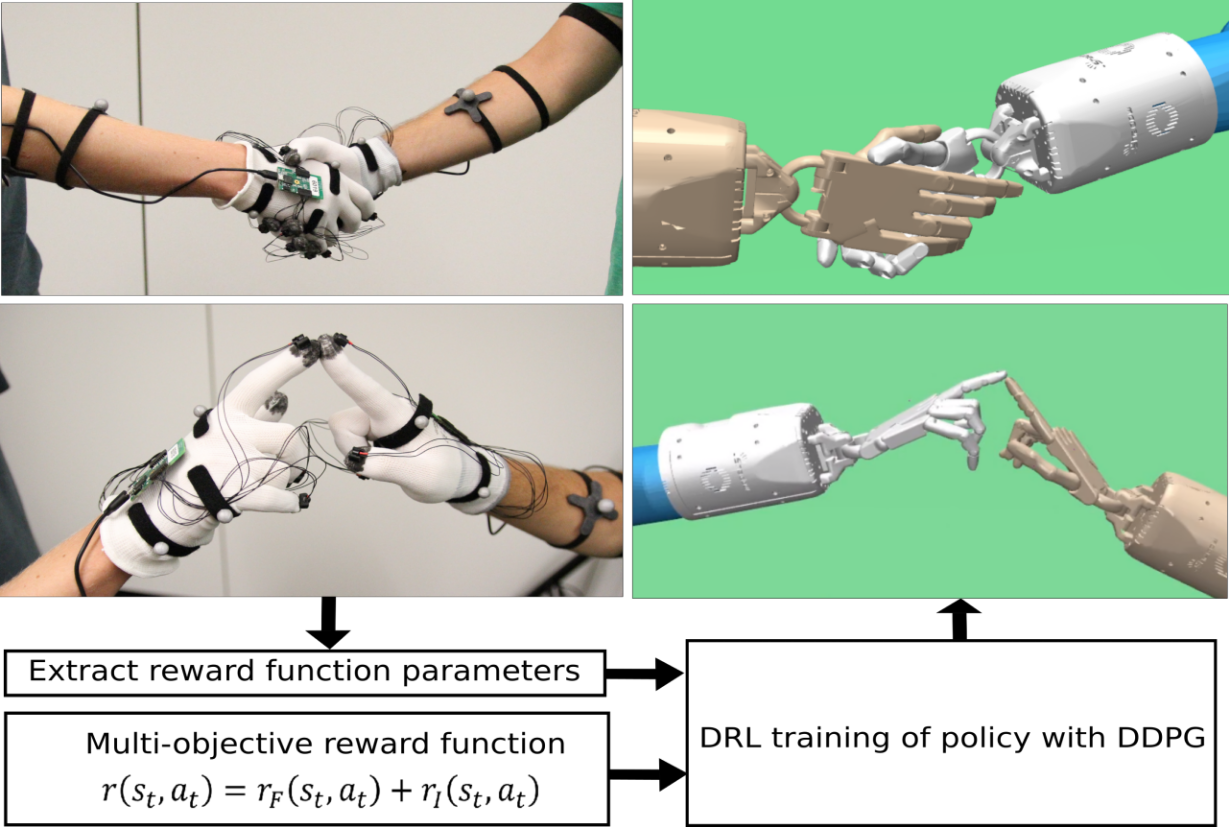}
	\caption{\small{Approach overview. The proposed multi-objective reward function and extracted parameters from human interaction data are used to train a human-robot interaction control policy via DRL.}}
	\label{fig:first_figure}
\end{figure}

This work presents a method for learning control policies for dexterous human-robot interaction. More specifically, we contribute the following:
\begin{inparaenum}[(i)]
	\item A multi-objective reward function for DRL algorithms. We show how reward function parameters are extracted from motion capture data and provide detailed analysis of how different parts of the reward influence the resulting control policy. 
	\item A training method which works in combination with standard DRL algorithms.
	\item A dataset of human hand interactions. 
	\item A large-scale user study showing that adding imitation reward to the policy results in motions that are perceived as more natural. 
\end{inparaenum}


\section{Related Work}

\subsection{Human Hand Interaction}

Different aspects of the human-robot handshake problem were investigated in the robotics community, e.g., force properties of a human handshake \cite{knoop2017handshakiness}, the possibility to recognize personality and gender from a handshake \cite{orefice2016let}, or the design of a compliant controller for handshakes \cite{arns2017design}.
Previous work mostly focuses on the handshake properties after the contact phase. However, producing a handshake movement is equally important \cite{shu2017learning, jindai2015handshake}. When humans establish a handshake, one person requests the handshake by holding out one hand, while the other person responds by grabbing the hand \cite{jindai2015handshake}. Based on this observation, \cite{jindai2015handshake} tries to model the appropriate time to request a handshake.



To achieve a handshake with humanoid hands, usually the robot requests the handshake and closes the fingers when the human hand is in contact \cite{arns2017design}. To the best of our knowledge, our method is the first that treats the problem in the case where the human requests the handshake. This case is harder from a control perspective, because it requires coordination with the human hand, the robot needs to produce natural looking motions and it still involves physical contact as in the previous case. The control of robotic arm movement is investigated in \cite{shu2017learning, jindai2015handshake}, but these papers do not control humanoid hands and do not observe complex contact dynamics. Furthermore, we investigate the possibility of performing different hand interactions, such as hand claps or finger touches, which are relatively under-researched.


\subsection{Control of Dexterous Humanoid Hands}

Dexterous humanoid hands are a highly complex mechanical systems \cite{kumar2013fast}. Due to their high complexity, the control problem is shown to be very challenging \cite{rajeswaran2017learning, mordatch2012contact}. Most approaches therefore use trajectory optimization to provide a controller \cite{mordatch2012contact, bai2014dexterous, kumar2014real}. These approaches require a precise model of the system, which makes them hard to transfer to real robots.
Leveraging real robot data for model learning was proposed in \cite{kumar2016optimal}, but the method is limited to slow in-hand manipulation of a pole. Contrary, model-free DRL does not require a model of the robot dynamics, i.e., all information is obtained through multiple episodes of trial-and-error. The only input to the algorithm is a reward function. Recently, model-free DRL has been applied to the control problem of humanoid hands \cite{rajeswaran2017learning},  achieving impressive results in a simulated environment. Furthermore, \cite{openAI2018} demonstrates the possibility of transferring a policy trained in simulation to the real Shadow Dexterous Hand. 


For non-linear control tasks, model-free DRL algorithms have shown impressive performance \cite{lillicrap2015continuous, schulman2015trust, schulman2017proximal} on the OpenAI gym benchmark problems \cite{openAiGym}. Furthermore, when applied to low degree of freedom robotic manipulators (7-10 DoF), like robotic hands with grippers, model-free DRL has been leveraged successfully \cite{gu2017deep, vecerik2017leveraging, rusu2017sim}. However, the problem becomes more challenging when DRL is applied to systems with higher DoF. A hand control problem \cite{rajeswaran2017learning} or natural movement character control problem \cite{peng2018deepmimic} deal with such systems. 
In \cite{peng2018deepmimic}, the authors carefully design and adjust the weights of the reward to achieve the desired performance. 
To improve convergence properties, the authors use two training techniques: setting the initial state on the demonstration trajectory and early stopping. 
Human demonstration can be used to accelerate the convergence rate \cite{rajeswaran2017learning}. In \cite{rajeswaran2017learning}, a human operator provides demonstrations via teleoperation, which are used to initialize the policy. 
Alternatively, IRL methods \cite{abbeel2004apprenticeship, ziebart2008maximum, wulfmeier2015maximum} can learn the reward function from demonstration data. However, they require running the RL algorithm in the inner loop of an iterative method, which is not feasible with DRL algorithms. To overcome this issue, IRL is posed as an adversarial imitation problem \cite{merel2017learning, ho2016generative}, but these methods are prone to instability.

Our method is inspired by previous DRL approaches, but our task requires significant changes of existing methods. In \cite{rajeswaran2017learning}, the demonstrations are provided by teleoperating a simulated robot hand, operating in isolation. 
For our case, it is impractical to collect demonstrations this way because it requires interactions between two humans. Thus, we capture real interactions using a motion capture system. 
Our method is more similar to \cite{peng2018deepmimic}, which is not designed for humanoid hands. 
In \cite{peng2018deepmimic}, the authors use motion capture data directly in the reward function formulation. Contrary, we extract the final pose parameters from data, while we similarly use motion capture data to produce natural looking motions. Additionally, we add contact patterns as an objective to the reward function, and provide a different training method.  

\section{Preliminaries}

\subsection{Deep Reinforcement Learning}
\label{sec:drl}

\newcommand{\StatesSet}{\mathcal{S}}
\newcommand{\ActionsSet}{\mathcal{A}}
\newcommand{\TransitionSet}{\mathcal{T}}
\newcommand{\state}[1]{{s}_{#1}}
\newcommand{\action}[1]{{a}_{#1}}
\newcommand{\reward}{r}
\newcommand{\discountRate}{\gamma}
\newcommand{\QFunction}{Q}
\newcommand{\policy}{\pi}
\newcommand{\Cost}{J}
\newcommand{\mdp}{\mathcal{M}}

Our control problem can be formalized using Markov Decision Processes (MDP), defined as a tuple $\mdp = \{\StatesSet, \ActionsSet, \mathcal{R}, \mathcal{T}, \rho_0, \discountRate \}$. We describe an environment with a set of states $\StatesSet$, a set of actions $\ActionsSet$, a reward function $\mathcal{R} = \reward(\state{t}, \action{t})$, transition dynamics $\mathcal{T} = p(\state{t+1}|\state{t}, \action{t})$, an initial state distribution $\rho_0 = p(\state{1})$ and a discount rate $\discountRate \in [0,1]$, where $\state{t} \in \StatesSet$ and $\action{t} \in \ActionsSet$. Model-free RL does not require knowledge of transition dynamics and requires only sampling from the transition dynamics probability distribution.  
We define the return $R_{t}$ as a discounted sum of future rewards:
\begin{equation}
R_{t} = \sum_{i=t}^{T}\discountRate^{i-t}\reward(\state{i}, \action{i}).
\end{equation}
The controller is defined as a control policy $\policy$, which maps states to actions $\policy: \StatesSet \rightarrow \ActionsSet$. The goal of the reinforcement learning algorithm is to learn a policy $\policy$ which maximizes the expected return from the start distribution:
\begin{equation}
\Cost = \mathbb{E}_{\mdp} \left [R_{1} \right ]
\end{equation}

We define the action-value or Q-function, which describes the expected return under a policy $\policy$ when taking action $\action{t}$ from state $\state{t}$, also called state-action pair, as follows: 
\begin{equation}
Q(\state{t}, \action{t}) = \mathbb{E}_{\mdp} \left [ R_{t}  \middle\vert \state{t}, \action{t} \right ].
\end{equation}

To solve the given problem, we define the Q-function and policy as neural network function approximators parametrized with $\theta^Q$ and $\theta^{\policy}$. This is known as an actor-critic type of RL algorithm, since we learn both an actor function, i.e., the policy, and a critic function, i.e., the Q-function. More specifically, we use the DDPG algorithm \cite{lillicrap2015continuous} to compute the gradients for updating the neural network parameters. To update  $\theta^Q$, we minimize the loss:
\begin{align}
&L(\theta^Q) = \mathbb{E}_{\mdp, \action{t}} \left [ \left( Q(\state{t}, \action{t} ; \theta^Q) - y_t \right)^2   \right] \\
&y_t = \reward(\state{t}, \action{t}) + \discountRate Q(\state{t+1}, \action{t+1} ; \theta^Q) 
\end{align} 
To update the actor parameters $\theta^{\policy}$, we compute the gradients:
\begin{equation}
\nabla_{\theta^{\policy}} \Cost = \mathbb{E}_{\mdp} \left [\nabla_{\theta^{\policy}} Q(\state{t}, \action{t} ; \theta^Q)    \vert \state{t}, \action{t} = \policy(\state{t} ; \theta^{\policy}) \right ],
\end{equation}
which are applied to the actor neural network. For both networks we use three fully connected layers with $256$ neurons and ReLu activation functions.

To ensure convergence of the policy, we apply all techniques from the DDPG paper to stabilize convergence properties. This includes a replay buffer, batch normalization and target networks. DDPG is an off-policy algorithm, thus we define the exploration policy as:
\begin{equation}
\action{t} = \policy(\state{t}) + \mathcal{N}, 
\end{equation}
where $\mathcal{N}$ is a sample from zero mean Normal distribution. More implementation details can be found in \cite{lillicrap2015continuous}.



\subsection{Simulation Environment}

\begin{figure}
\centering
	\includegraphics[scale=0.17]{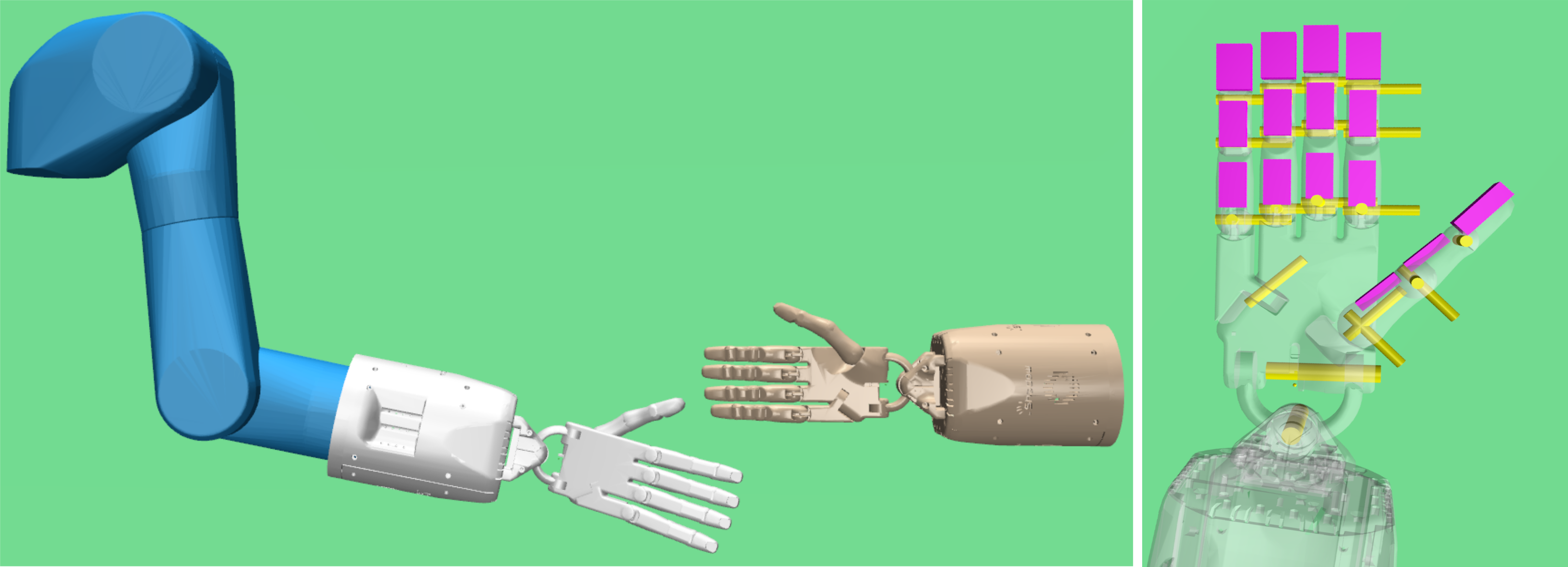}
	\caption{\small{Agent arm with the Shadow Dexterous Hand. The system has 28 DoF. The arm, shown in blue in the left figure, has 4 Dof. The hand has 24 Dof, depicted as yellow cylinders in the right figure, and 20 actuators. Contact sensors are marked in purple (right).}}
	\label{fig:dof}
\end{figure}

Our simulation environment consist of two robots: the agent, controlled by the policy, and the target hand. The agent consists of a 4 DoF robotic arm and the Shadow Dexterous Hand with 24 DoF. The Shadow Dexterous Hand is controlled by 20 actuators (cf. Fig.~\ref{fig:dof}). We use the same hand model as stand-in for a human hand for convenience, since this model is easy to pose in different configurations. However, this model can be replaced with a human hand model, which should not influence the results of our experiments since the hand is not actuated, as usually assumed for a hand requesting an interaction \cite{shu2017learning, jindai2015handshake}. The robot models are taken from the OpenAI gym framework \cite{openAiGym}. 

The input to the control policy are joint angles, joint velocities and contact sensor readings (cf. Fig.~\ref{fig:dof}) of the agent hand. Additional inputs are the positions of the target hand links and the origin of each rigid body on the hand. The policy outputs are control signals that actuate the agent's arm and hand. Control signals are setpoints for the joint angles scaled in the range from $-1.0$ to $1.0$. 

\section{Method}

Our method is able to learn control policies of hand interactions using motion capture data of human demonstrations. We assume the following setting: the first participant requests the interaction while the second is executing the interaction sequence. In the example of a handshake, the first participant stretches out the hand to request the handshake, while the other responds by grabbing the hand. The robot learns to perform the behavior of the second participant. The goal is to produce the desired interaction \emph{and} motions perceived as natural. 
We propose a \emph{single} reward function that can be applied to various hand interactions. 
The parameters of the reward function are extracted directly from the motion capture dataset using Alg.~\ref{alg:1}. The policy is trained via a DDPG based training method, as explained in Sec.~\ref{sec:training_alg}.



\newcommand{\gain}{\omega}
\newcommand{\dataset}{\mathcal{D}}
\newcommand{\indicator}{\mathds{1}^i_{ct}}

\subsection{Reward Function}
\label{sec:rew_func}

Our proposed reward function consists of two terms:
\begin{equation}
\reward(\state{t}, \action{t}) = \reward_F(\state{t}, \action{t}) + \reward_I(\state{t}, \action{t}),
\end{equation}
where $\reward_F(\state{t}, \action{t})$ is the final state reward, which is used to reward the correct end configuration, and  the imitation state reward $\reward_I(\state{t}, \action{t})$, which provides trajectory guidance to make the interactions look more natural.
The final state reward itself consists of four terms:
\begin{equation}
\reward_F(\state{t}, \action{t}) = \reward_p(\state{t}) + \reward_{\alpha}(\state{t}) + \reward_c(\state{t}) + \reward_a(\action{t}),
\end{equation}
where $\reward_p(\state{t})$ is a position reward, $\reward_\alpha(\state{t})$ is an angle reward, $\reward_c(\state{t})$ is a contact reward, and $\reward_a(\action{t})$ penalizes high action inputs. Experimentally, we determined that the most important position features are the fingertip positions of the agent hand (total number $N_f = 5$). Regarding the angles, we use all joint angles of the robot hand to compute the angle reward (total number $N_{\alpha} = 24$). Position and angle rewards are defined as a negative $l_2$ norm of position features and angle errors:
\begin{align}
& \reward_p(\state{t}) = - \sum_{i=1}^{N_f} \omega^i_p  \vert\vert p^i_g - p^i_{rt} \vert \vert \\
& \reward_{\alpha}(\state{t}) = - \sum_{i=1}^{N_{\alpha}} \omega^i_{\alpha} \vert\vert \alpha^i_g - \alpha^i_{rt} \vert\vert.
\end{align}
The vector $p^i_g$ is the goal position of each position feature and $p^i_{rt}$ is the current position of the respective feature. Similarly, $\alpha^i_g$ is the goal joint angle and $\alpha^i_{rt}$ the current joint angle on the robot hand. The weights $\omega^i_p$, $\omega^i_{\alpha}$ determine the importance of the specific goal, which we define via algorithm described in Sec.~\ref{sec:reward_alg}. When using only position and angle rewards, the robot hand only roughly reaches the desired end configuration (cf. Fig~\ref{fig:experiment_1_poses}, Baseline 2). 
Our task requires accurate hand positioning, which is hard to achieve in tasks that involve contacts.
We achieve the desired hand position by adding a contact reward $\reward_c(\state{t})$, which forces the desired contact profile.
For more details about the influence of the reward terms we refer to Sec.~\ref{sec:abl_study}.
Additionally, a control input reward is added to prevent high control signals. Contact and input rewards are defined as:
\begin{align}
& \reward_c(\state{t}) = \sum_{i=1}^{N_{c}} \gain^i_c \indicator \quad ,\\
& \reward_a(\action{t}) = - \sum_{i=1}^{N_{a}} \gain^i_a  \vert\vert a^i \vert\vert^2 \quad ,
\end{align}
where the indicator function $\indicator$ outputs $1$ in case the contact sensor is active and $0$ otherwise. The weight $\gain^i_c$ determines the importance of each contact sensor. The system has $N_{c} = 15$ contact sensors. $a^i$ is the control input signal for each actuator and $\gain^i_a$ is the respective weight. The system has $N_a = 24$ control inputs.


The imitation reward consist of two further terms:
\begin{equation}
\reward_I(\state{t}, \action{t}) = \reward_{p I}(\state{t}) + \reward_{\alpha I}(\state{t}).
\end{equation}
The difference compared to the final state reward is that the goal position $p^i_{g t}$ and goal angles $\alpha^i_{g t}$ depend on the timestep:
\begin{align}
& \reward_{p I}(\state{t}) = - K_{p I} \sum_{i=1}^{N_f} \omega^i_{p} \vert \vert p^i_{g t} - p^i_r \vert \vert \quad ,\\
& \reward_{\alpha I}(\state{t}) = - K_{\alpha I} \sum_{i=1}^{N_{\alpha}} \omega^i_{\alpha} \vert\vert \alpha^i_{gt} - \alpha^i_r \vert\vert.
\end{align}
These rewards are scaled with the weights $K_{p I}$ and $K_{\alpha I}$.


The final state reward function is often enough to complete the interaction, i.e., to reach the final pose. However, the most direct trajectory often does not look natural (cf. Fig.~\ref{fig:experiment_2_qual_clap}). 
We overcome this issue by adding an imitation reward. As shown in Sec.~\ref{sec:imitation_eval}, imitation reward significantly improves that the hand motion is perceived as natural.
However, when the starting hand pose is far away from poses in demonstration examples, the policy may produce non-smooth motions. We evaluate these examples also in Sec.~\ref{sec:imitation_eval}.

\subsubsection{Reward function parameters}
\label{sec:reward_alg}
To define the terms of the reward function, we use a  motion capture dataset $\dataset = \{(p^{i}_{t}, p^{j}_{t}, \alpha^{i}_t, \alpha^{j}_t) \}_{t=1}^T$ of an interaction sequence. The dataset provides positions of rigid bodies $p^{i}_{t}, p^{j}_{t}$ and joint angles $\alpha^{i}_t, \alpha^{j}_t$ of two human hand models at timestep $t$: the target hand, denoted with superscript $j$, which requests the interaction and the actor hand, denoted with superscript $i$, which the robot imitates. From $\dataset$, we can calculate distances between the rigid bodies $d^{ij}_t$ and the relative position calculated in the coordinate frame of the target hand body $\Delta  p^{ij}_t$. 
We use fingertip positions on both hands, plus the palm position on the target hand. For the joint angles, we use all joints from the human data that have a corresponding joint on the robot hand. Based on the minimum distance, we set the reference frame $j_{\min}$ and reference timestep $t_{\min}$, as shown in Alg.~\ref{alg:1}.  Using these references, we compute the goal positions and goal angles. 
The position goals are defined in a goal centric way, which enables us to calculate them for a randomly positioned hand in the simulation $ p^j_{s}$, as shown in line 5 of Alg.~\ref{alg:1}, where $R^j_s$ is the rotation matrix of the target hand rigid body with index $j$.
\begin{algorithm}
\caption{Reward Parameters}\label{alg:1}
\begin{algorithmic}[1]
\State \textbf{Input:} $\dataset, d^{ij}_t, \Delta  p^{ij}_t$ 
\State $t_{\min} \gets \arg\min d^{ij}_t$, \quad $j_{\min} \gets \arg\min d^{ij}_t$
\State $p_s^{j_{\min}} \gets$ position of a $j_{\min}$ link in simulation 
\State $R_s^{j_{\min}} \gets$ rotation matrix of a $j_{\min}$ link in simulation
\State $p^i_g \gets p^{j_{\min}}_s + R^{j_{\min}}_s \Delta p_{t_{\min}}^{ij_{\min}}$,  \quad $\alpha^i_g \gets \alpha_{t_{\min}}^i$
\State $p_{g t}^i \gets p_s^{j_{\min}} + R_s^{j_{\min}} \Delta p_{t}^{ij_{\min}}$,  \quad $\alpha_{g t}^i \gets \alpha_{t}^i$
\State \Return $p_g^i,\alpha_g^i, \{\alpha_{g t}^i \vert t=1..t_{\min} \}, \{p_{g t}^i \vert t=1..t_{\min} \} $
\end{algorithmic}
\end{algorithm}


Position weights are calculated using the equation:
\begin{equation}
\omega^i_p = K_p \frac{d_{\min}}{d_{t_{\min}}^{ij_{\min}}}.
\end{equation}
The angle and control weights all have the same value $\omega_{\alpha}^i = K_{\alpha}, ~ \gain^i_a = K_{a}$.
We set the contact weights $\gain^i_c$ to $K_c$ for all sensors that should be in contact. This can be done by asking participants where they feel the pressure during interactions. Alternatively, one could use the method from \cite{knoop2017handshakiness}, which simply applies color to the target hand and measures contact area from paint marks. 

To train the policy, we need to set just six weights in the reward function ($K_p, K_{\alpha}, K_c, K_{a}, K_{p I}, K_{\alpha I} $). In all our experiments, $K_{a}$ is set to 1, while $K_c$ is roughly set to $\frac{N_{o}}{5}$, where $N_{o}$ is the number of sensors that should be in contact.

\subsection{Training} 
\label{sec:training_alg}

To train the policy, we first need to position the target hand. For a single training episode, the target hand stays fixed. Although we train the policy with a static hand, we experimentally show that our policy generalizes to moving hands (cf. \ref{sec:policy_eval}). The joint angles of the target hand are set according to the joints of the human target hand at a timestep which occurs prior to the interaction timestep $t_{\min}$. This ensures that the target hand is not closed, thus allowing the robot to interact. Our reward function is defined in a goal centric way. This enables randomization of the target hand position, performed at the beginning of each episode. 
To calculate the reward function parameters, we pick an interaction sequence uniformly at random.

We randomize the robot hand position additionally to target hand randomization. For imitation reward to be effective, the robot hand should be positioned in the same configuration as the human hand at the start of the imitation trajectory. Thus, we position the robot wrist to the position of the human wrist at timestep $t_s$, augmented with random Gaussian noise. The timestep $t_s$ is selected uniformly at random from a set $\{t_k \vert k=1 .. (t_{\min} - t_{\mathrm{off}})\}$. We use a small offset $t_{\mathrm{off}}$ because hands can collide in the last part of the trajectory. If this position is not reachable, we start from the closest reachable position. Contrary to \cite{peng2018deepmimic}, we cannot position the agent exactly in the human pose because of the configuration differences of the human and robot arms. Furthermore, our task is driven by a goal pose, which means that starting only from demonstration trajectories, as in \cite{peng2018deepmimic}, will result in poor generalization. After each $N_e$ steps, we update the network using the DRL algorithm described in Sec.~\ref{sec:drl}.

\subsection{Data Collection}

\begin{figure}
\centering
	\includegraphics[width= 0.60 \linewidth]{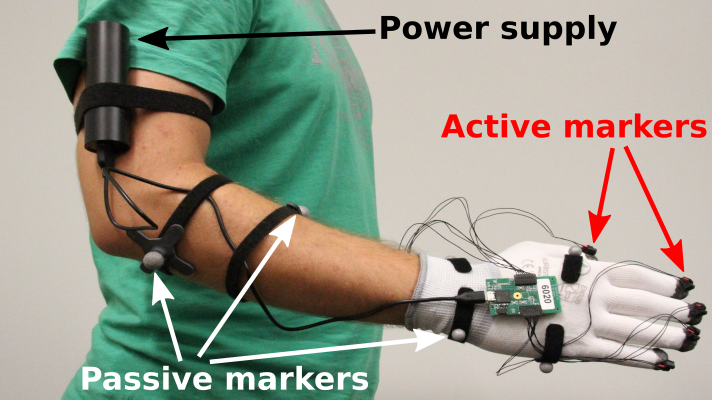}
	\caption{\small{Participants wear 5 active markers on the fingertips and 6 passive markers on the palm and forearm of the right hand.}}
	\label{fig:system}
\end{figure}

To collect data, we use the OptiTrack motion capture system. Each participant is equipped with markers as shown in Fig. \ref{fig:system}. We only track the right hand of each participant. 
Hand tracking is prone to marker mislabeling, with fingertip markers being most problematic. We use active markers, which can be uniquely identified by their blinking pattern, on the fingertips.
The OptriTrack software fits a model of the human hand to the markers, providing the position of each link and joint angle of the human hand. For each interaction, we recorded five demonstrations. We will release the dataset and simulation environment for further research (\href{https://ait.ethz.ch/projects/2019/DRL-handshake/}{https://ait.ethz.ch/projects/2019/DRL-handshake/}).



\section{Results}

We conducted experiments in simulation to evaluate our method. We extensively test our policy on three different hand interactions: handshake, hand clap, and E.T. greeting (e.g. index finger touching), see Fig. \ref{fig:experiment_1_poses}. These three interactions are diverse: the handshake requires grasping of the target hand in a specific way, the hand clap has a characteristic motion prior to contact, while the E.T. greeting requires precise positioning of the index finger. 

\subsection{Ablation Study on Reward Function}
\label{sec:abl_study}

\begin{figure}
\centering
	\includegraphics[width=.95\linewidth]{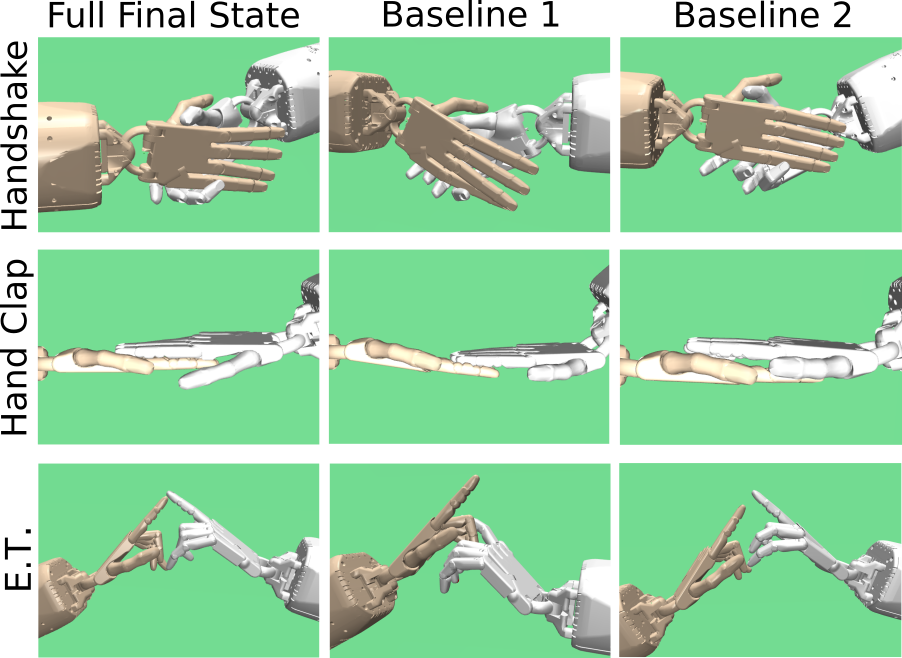}
	\caption{\small{Final poses of the hands for different hand interactions.}}
	\label{fig:experiment_1_poses}
\end{figure}

In a first experiment, we intend to show the influence of the different parts of our reward function on the resulting control policy.
For this, we do an ablation study on our reward function.
We compare the full final state reward $\reward_F(\state{t}, \action{t})$ with two baselines. Baseline 1 uses only the relative position of the palm instead of the fingertips as a goal position feature, while keeping angle, contact and input reward the same. In Baseline 2, we remove the contact reward from the reward function. Furthermore, we examine the influence of adding the imitation reward to the final state reward.  


\begin{figure*}
\centering
	\includegraphics[width=0.98\linewidth]{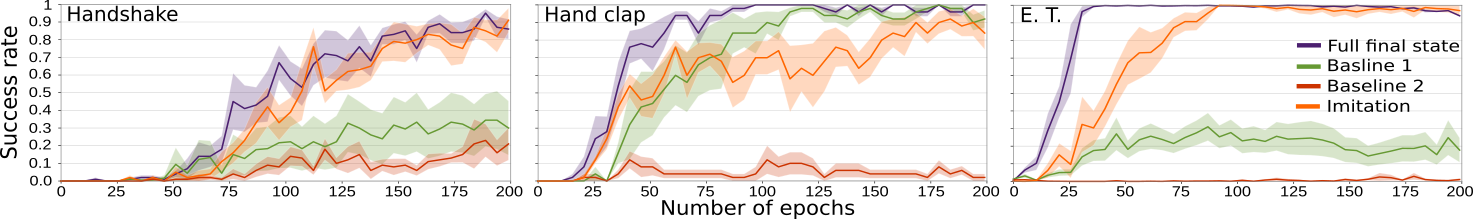}
	\vspace{-0.2cm}
	\caption{\small{Baseline comparison. We measure the success rate of the policy every five epochs. Success is estimated from the contact profile on the hand. We show average results from 5 random seeds with the standard error indicated by the shaded area.}}
	\label{fig:experiment_1_fig}
	\vspace{-0.3cm}
\end{figure*}

Our final state reward function shows overall better performance than both baselines (c.f. Fig.~\ref{fig:experiment_1_fig}). The influence of the position reward can be seen by comparing Baseline 2 to Baseline 1 for handshake and E.T. interactions. Although both baselines have low success rates, Baseline 2 results in final configurations closer to the desired ones as shown in Fig.~\ref{fig:experiment_1_poses}. Adding contact reward to Baseline 2, i.e. using our reward function, removes these errors. The importance of the contact reward can be also seen in Fig.~\ref{fig:experiment_1_fig} in case of the hand clap. Since precise positioning is less important here, Baseline 1 achieves high success rate because of the contact reward. After adding the imitation reward, we observe that the success rates do not significantly change.

\subsection{Evaluation of Imitation Training}
\label{sec:imitation_eval}
To qualitatively assess the impact of the imitation reward, we conduct a large-scale user study ($N=116$). We present 11 video sequences of policy outputs with and without imitation reward side-by-side . We keep the initial conditions for each sequence-pair the same and randomly assign videos to the left or right. The participants state which video is perceived as more natural on a forced alternative choice 5-point scale. The five responses are: "Left sequence looks much more natural", "Left sequence looks more natural", "Both the same", "Right sequence looks more natural", "Right sequence looks much more natural".



\begin{figure}
\centering
	\includegraphics[width=.95\linewidth]{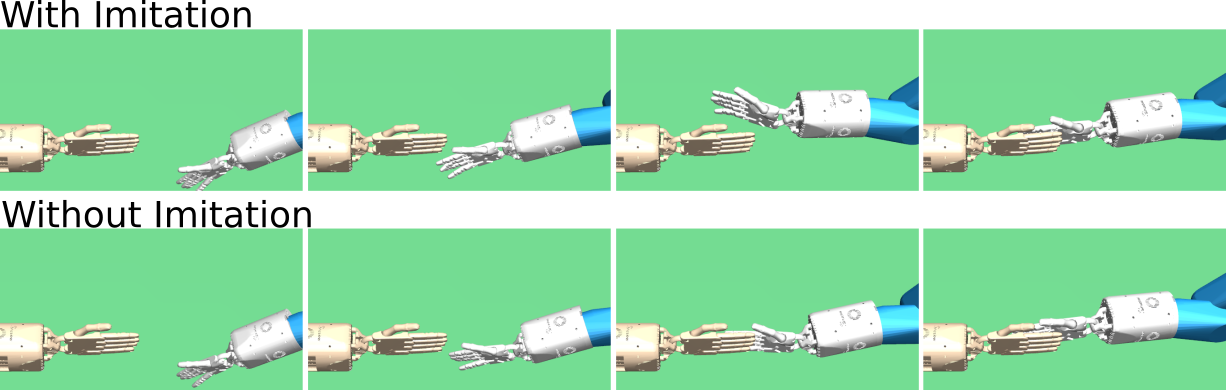}
	\caption{\small{Imitation reward. The policy trained with imitation reward produces a characteristic motion prior to the hand clap.}}
	\label{fig:experiment_2_qual_clap}
\end{figure}


Assuming equidistant intervals, we mapped user responses onto a scale from -2 to 2, where positive values mean that the user prefers the policy generated with imitation reward and vice versa. The imitation policy can generate non-smooth motions when the starting pose is far away from the recorded human trajectory. To evaluate these examples, we compare two sequences including obvious non-smooth motions.

\begin{table}
\centering
\caption{User Study results of people voting on 5 point scale from -2 (no imitation) to 2 (imitation) looking much more natural}
\vspace{-6pt}
\resizebox{\columnwidth}{!}{
\begin{tabular}{| c | c | c | c | c | c | c |}
\hline
Score & $-2$ & $-1$ & $0$ & $1$ & $2$ & mean \\
\hline
Handshake & $4.9\%$ & $26.2\%$ & $15.2\%$ & $\mathbf{45.1\%}$ & $8.6\%$ & $0.26$ \\
\hline
Hand clap & $0.0\%$ & $2.3\%$ & $3.4\%$ & $41.9\%$ & $\mathbf{52.3\%}$ & $1.44$ \\
\hline
E.T. & $3.4\%$ & $17.2\%$ & $\mathbf{54.3\%}$ & $19.0\%$ & $6.0\%$ & $0.07$ \\
\hline
Handshake non-smooth & $23.3\%$ & $\mathbf{43.1\%}$ & $12.9\%$ & $15.5\%$ & $5.1\%$ & $-0.64$ \\
\hline
Hand clap non-smooth & $0.8\%$ & $7.0\%$ & $12.9\%$ & $\mathbf{49.1\%}$ & $30.2\%$ & $1.0$ \\
\hline
\end{tabular}\label{tbl:policies_compare}
}
\vspace{-14pt}
\end{table}

Generally speaking, participants favor policies generated with imitation reward (c.f. Table \ref{tbl:policies_compare}). For hand claps, the differences are easy to spot (see Fig.~\ref{fig:experiment_2_qual_clap}). Hence, human raters strongly prefer the imitation based policy. For the handshake, the differences are harder to see, resulting in significant amount of participants selecting "Both the same" ($15.2\%$). However, the majority of the participants still prefer the imitation based policy. 
For the E.T. interactions, there are no observable differences. Thus, the majority of participants answered with "Both the same" ($54.3\%$).
For non-smooth handshakes, the results indicate that participants prefer smooth motions. However, for non-smooth hand claps, participants prefer imitation features, although the mean score is lower than in the case of smooth hand claps.

\subsection{Policy Evaluations}
\label{sec:policy_eval}

\begin{figure}
\centering
	\includegraphics[width=.95\linewidth]{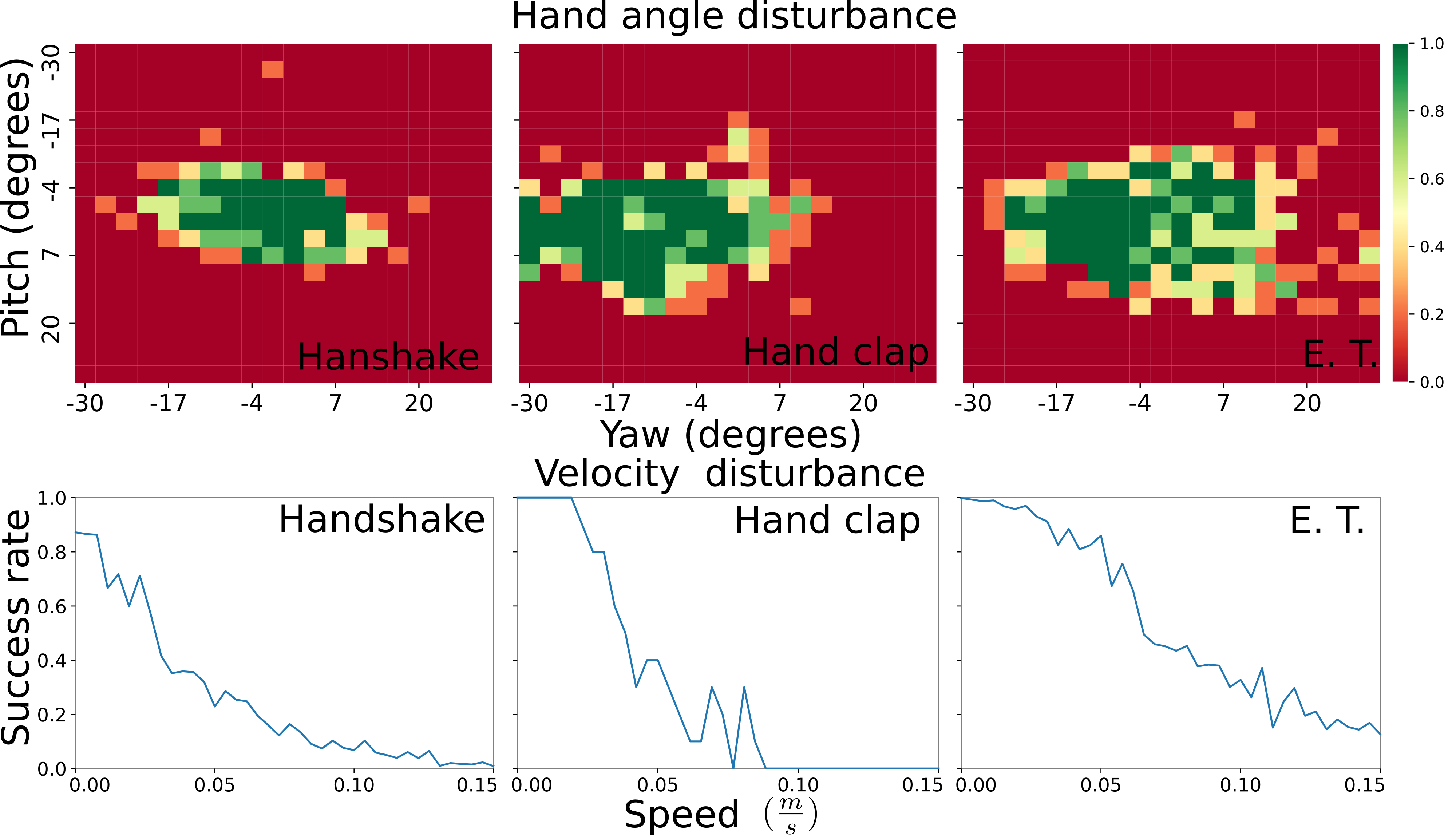}
	\caption{\small{Robustness experiments. Top: The average success rate for different target hand angles (the color bar shows the success rate). Down: The average success rate for different hand velocities. }}
	\label{fig:robustnes}
\end{figure}

To evaluate the robustness of policies created with our method, we test the reaction to perturbations in orientation and velocity of the target hand. During training, we only randomize the position of the target hand, while the orientation stays the same. We position the target hand in the reachable workspace of the agent and measure the success rate, while changing the yaw and pitch angles of the target hand. In a realistic scenario, the human hand will not be perfectly still. Furthermore, the human can react when the contact is imminent by closing the hand or approaching the agent hand. Thus, we conduct a second experiment where the target hand is moving at constant speed, changing the direction at random every half second (see Fig.~\ref{fig:robustnes}).

The experiments indicate that our method is robust to perturbations in orientation and velocity of the target hand. This shows that our reward function generates policies that generalize well to unseen scenarios. 
We also tested our policy with changing configurations of the target hand, i.e., closing fingers in a handshake policy, and did not observe any major changes. Demonstrations of these experiments can be seen in the accompanying video (\href{https://youtu.be/ZSgEqyltaN4}{https://youtu.be/ZSgEqyltaN4}).


\section{Discussion and Conclusion}

Control of dexterous humanoid hands is a challenging problem, especially when it involves contact dynamics. 
In this paper, we demonstrate that a single parametrized reward function can be used for different hand interactions. 
To define parameters of the reward function, we use a simple algorithm to extract parameters from motion capture data. 
We show that policies generated with our method produce more natural looking trajectories, and generalize well to different orientations and velocities of the target hand.

Our results are shown only in simulation and with a static target hand as an initial step towards natural human-robot hand interactions. To achieve this level of performance on a real robot, transfer learning methods, such as the one suggested in \cite{openAI2018}, could be applied. 
We show that our policy reacts well to small velocity disturbances. However, humans can perform synchronous hand motions prior to interaction. This problem should be investigated in more detail. Our method only considers contacts, but we never investigated forces acting on the hand. \cite{knoop2017handshakiness} emphasizes the importance of forces applied during handshakes. According to our measurements, forces applied to the target hand are in the range of a normal handshake. 
Compliant behavior is important for hand interactions \cite{arns2017design}. However, evaluation of the robot hand compliance  is outside the scope of our work.    


This paper shows how natural human-robot hand interaction can be learned using DRL. To the best of our knowledge, this is the first paper that uses a dexterous humanoid hand for human-robot hand interactions. This opens up the possibility to achieve natural hand interactions on a real humanoid robot.






\bibliographystyle{IEEEtran}

\bibliography{IEEEabrv,references}




\end{document}